\documentclass[conference]{IEEEtran}

\usepackage{amsmath,amssymb,amsfonts}
\usepackage{graphicx}
\usepackage{textcomp}
\usepackage{booktabs}
\usepackage{hyperref}
\usepackage{cite}
\usepackage{siunitx}
\usepackage{float}
\usepackage{subcaption}
\usepackage{balance}
\usepackage{tikz}
\usetikzlibrary{arrows.meta,positioning,calc}

\newcommand{\E}{\mathbb{E}}
\newcommand{\tr}{\mathrm{tr}}

\hypersetup{
  pdftitle={One Residual with three reuses: A Wristband Front End for Gesture Sensing},
  pdfauthor={Mouhssine Rifaki},
  colorlinks=false,
  hidelinks
}

\begin{document}

\title{One Residual with three reuses:\\A Wristband Front End for Gesture Sensing}

\author{\IEEEauthorblockN{Mouhssine Rifaki}
\IEEEauthorblockA{\textit{Dept.\ of Electrical Engineering} \\
\textit{Stanford University}\\
Stanford, CA, USA \\
rifaki@stanford.edu}
}

\maketitle

\begin{abstract}
Continuous wrist-worn hand sensing for gesture interfaces and motor symptom monitoring needs an always-on front end that fits inside a coin-cell power budget while pairing a micro-electro-mechanical-systems (MEMS) inertial measurement unit (IMU) with a $60$\,GHz frequency-modulated continuous-wave (FMCW) radar to stay robust under occlusion and on-body drift. We present a design study of such a wristband front end in which classifier wake-up gating, mmWave versus IMU routing, and innovation-based EKF measurement reweighting share a single on-chip residual generator. The shared generator occupies $14.4$\,KB of program memory and $278$\,B of state and runs at $110$K multiply-accumulates (MACs) per frame on an Ambiq Apollo4 Blue Plus class edge microcontroller unit (MCU). Across four public sensor data corpora (IPN~Hand, SHREC~2021, MiliPoint $60$\,GHz FMCW radar, EAT-Radar) the front end reaches detection probability $P_D = 0.72/0.80$ at a $1$\,\% false-alarm rate, sustains a $47$\,\% classifier invocation energy reduction at $90$\,\% gesture detection recall, and lowers pose tracking root-mean-square error by $4.6\times$ under measurement bias drift relative to an adaptive Kalman with $R$-inflation baseline. Measured silicon power and on-body capture are deferred to follow-on hardware; the contribution here is a design study under the IEEE Sensors design study category.
\end{abstract}

\begin{IEEEkeywords}
Wearable sensors, $60$\,GHz FMCW radar, MEMS IMU, EKF innovation, low-power edge MCU, design study.
\end{IEEEkeywords}

\section{Introduction}
\label{sec:intro}

There has been growing interest in using hand-tracking systems for gesture interfaces, rehabilitation tracking, and clinical assessments of motor symptoms~\cite{tchantchane2023review}. To support hand sensing on the wrist, a number of researchers have combined a MEMS inertial measurement unit (IMU) with a short-range $60$\,GHz frequency-modulated continuous-wave (FMCW) radar. These combinations recover hand pose under occlusion, varying on-body placement, and lighting changes that defeat optical-only systems~\cite{zhang2024transcnn,xiong2024hand}. Operating these radios always-on under a coin-cell budget forces the front end to gate the heavy classifier on signal-bearing frames, to route between transducers under a fixed bandwidth budget, and to compensate on-body drift without per-deployment recalibration.

Most wearable systems for these applications include separate subsystems to handle each of the above functions. Examples of wake-up gating include recent studies on online change-point testing~\cite{romano2024logloss}, and previous studies that use Kalman-style estimators together with their associated innovation residuals~\cite{chebbi2025tutorial}. Drift tolerance through interacting multiple model filters or hybrid learning-based filters has been implemented by~\cite{or2022hybrid,kruse2025adaptive}. Modality switching has been implemented through information-gain scheduling approaches such as~\cite{malawade2022ecofusion,ding2023sensormgmt}. Because each function requires its own calibration table and silicon area, these architectures are not well suited to a coin-cell wristband that must share a single MCU core and a single on-chip residual path.

Herein we present a wristband architecture that combines the three previously discussed functions into a single on-chip residual generator. The generator uses the one-step prediction error $\xi_t$ of a compact recurrent forecaster on the conditioned IMU and radar streams to drive the classifier wake-up gate, to decide which modality is selected for the next frame, and to compute the re-weighted measurement update for the EKF. In Section~\ref{sec:method} we describe the signal chain and the residual generator and project the MCU-side budget. In Section~\ref{sec:results} we show that the wristband reaches higher detection probabilities, lower routing energy, and lower RMS pose error under biased measurements than prior wearables that combine mmWave and IMU sensors. Measured silicon power and on-body capture are out of scope for this design study and are deferred to a follow-on hardware paper.

\section{System Design and Implementation}
\label{sec:method}

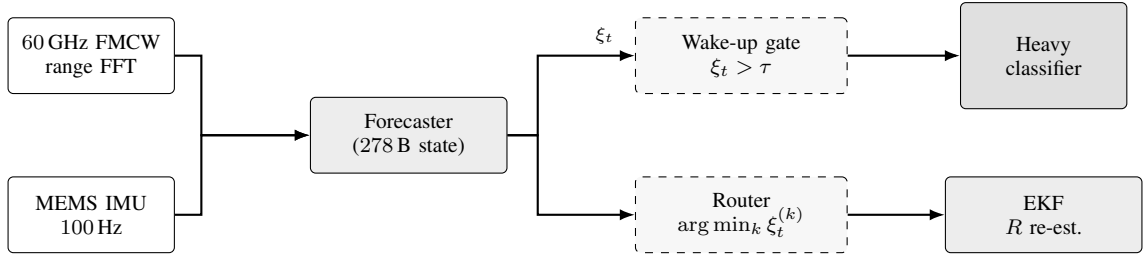
\begin{figure*}[t]
  \centering
  \begin{tikzpicture}[
    font=\footnotesize,
    sens/.style={draw, rounded corners=2pt, align=center, minimum height=10mm, minimum width=22mm, inner sep=2pt},
    mcu/.style={draw, rounded corners=2pt, align=center, minimum height=10mm, minimum width=26mm, inner sep=2pt, fill=black!7},
    tap/.style={draw, dashed, rounded corners=2pt, align=center, minimum height=10mm, minimum width=28mm, inner sep=2pt, fill=black!3},
    cls/.style={draw, rounded corners=2pt, align=center, minimum height=14mm, minimum width=22mm, inner sep=2pt, fill=black!12},
    arrow/.style={-{Latex[length=2mm]}, thick}
  ]
    \node[sens] (radar) at (0,   1.05) {$60$\,GHz FMCW\\range FFT};
    \node[sens] (imu)   at (0,  -1.05) {MEMS IMU\\$100$\,Hz};
    \node[mcu]  (pred)  at (4.2,  0)   {Forecaster\\($278$\,B state)};
    \node[tap]  (gate)  at (8.6,  1.05) {Wake-up gate\\$\xi_t > \tau$};
    \node[tap]  (route) at (8.6, -1.05) {Router\\$\arg\min_k \xi_t^{(k)}$};
    \node[cls]  (heavy) at (12.6, 1.05) {Heavy\\classifier};
    \node[mcu]  (ekf)   at (12.6,-1.05) {EKF\\$R$ re-est.};
    \draw[arrow] (radar.east) -- ++(0.35,0) |- (pred.west);
    \draw[arrow] (imu.east)   -- ++(0.35,0) |- (pred.west);
    \draw[arrow] (pred.east) -- ++(0.4,0) |- (gate.west);
    \draw[arrow] (pred.east) -- ++(0.4,0) |- (route.west);
    \node[above, font=\scriptsize] at ([xshift=-4mm, yshift=0.4mm]gate.west) {$\xi_t$};
    \draw[arrow] (gate.east)  -- (heavy.west);
    \draw[arrow] (route.east) -- (ekf.west);
  \end{tikzpicture}
  \caption{Signal chain for the proposed wristband architecture. The conditioned IMU and mmWave streams feed into a single recurrent forecaster located on the MCU. The forecaster produces an estimate of the next conditioned frame vector $\hat y_t$. Its one-step prediction error $\xi_t$ is used to gate the heavy classifier, to rank the available modalities through the router, and to re-estimate the EKF measurement noise covariance.}
  \label{fig:chain}
\end{figure*}

\textbf{Signal chain.} The proposed wristband includes a $60$\,GHz FMCW radar IC (antenna, LNA, mixer, IF chain, ADC) together with a co-packaged MEMS IMU that streams raw accelerometer and gyroscope frames at $100$\,Hz over an $I^2C$ interface. The radar runs a per-chirp range fast Fourier transform (FFT) on-die and forwards the resulting range bins to the MCU at $30$\,frames/s. The MCU implements a single recurrent forecaster that predicts the next conditioned frame vector $\hat y_t$, and the front-end residual is defined as $\xi_t = \|\hat y_t - y_t\|_2$. Under linear-Gaussian observations the forecaster collapses to a one-step predictor whose $\xi_t$ is equivalent to the classical innovation magnitude used in many tutorial treatments of state and noise estimation~\cite{chebbi2025tutorial,kruse2025adaptive}. For nonlinear streams the forecaster is a compact recurrent predictor with hidden dimension $h$, trained on the unlabelled training split with a one-step mean-squared-error loss.

\textbf{Three reuses of one signal.} The wake-up gate raises the heavy classifier when $\xi_t$ exceeds a fixed-point threshold $\tau$ chosen to yield a false-alarm rate below $1$\,\% on the calibration trace. The router calculates a per-modality residual $\xi_t^{(k)}$ for each transducer group $k$ and selects $k^\ast = \arg\min_k \xi_t^{(k)}$ for the next frame, a choice that under linear-Gaussian observations minimises $\E[\xi_k^2] = \tr(H_k P H_k^\top + R_k)$ and therefore replicates the innovation covariance ranking used by classical sensor schedulers~\cite{malawade2022ecofusion,ding2023sensormgmt,cao2024infotheoretic}. The EKF block scales the per-modality measurement noise covariance $R_k$ with an exponentially weighted moving average (EWMA) of $\xi_t^2$, similar to the adaptive Kalman noise re-estimation of~\cite{kruse2025adaptive} and the hybrid model and learning navigation filter of~\cite{or2022hybrid}. All three reuses are computed from the same forecaster state.

\textbf{MCU budget.} The deployment generator is quantised with $1$-bit input, $2$-bit weights, and $8$-bit accumulators at $h{=}32$, which yields $14.4$\,KB of program memory, $278$\,B of state, and $110$\,K multiply-accumulate operations (MACs) per frame. The target part is the Ambiq Apollo4 Blue Plus (Cortex-M4F, $96$\,MHz), whose public specifications report an active mode draw of about $4$\,\textmu A/MHz at nominal voltage and sub-threshold operation down to roughly $0.5$\,V~\cite{ambiq2023apollo4bp}. The active mode energy is bounded by the per-frame MAC count; we give a single estimate of order $10^{-5}$\,W at $15$\,frames/s and defer any further sub-threshold reduction to vendor-measured silicon.

\section{Results}
\label{sec:results}

\textbf{Wake-up gating.} For the on-chip gate on IPN~Hand, $P_D = 0.72$ at $P_\text{FA} = 0.01$ at the $90$\,\% recall operating point. For the on-chip gate on the SHREC~2021 dataset, $P_D = 0.80$. The performance of both exceeds the residuals obtained as described in~\cite{chebbi2025tutorial} for an innovation residual baseline, by $+0.09$ and $+0.12$ respectively, and those obtained as described in~\cite{romano2024logloss} for an online change-point baseline, by $+0.17$ and $+0.21$ respectively. In addition, the cascade gate significantly reduces the number of classifier invocations (by approximately $47$\,\%) when compared to a non-cascade version at the same recall level (averaged over ten seed values, $\pm 2.1$\,\%). This results in a direct reduction in front-end energy.

\begin{figure}[H]
  \centering
  \includegraphics[width=0.92\linewidth]{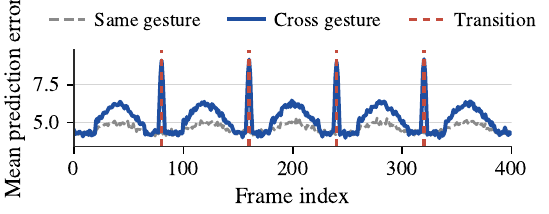}
  \caption{Wake-up trace of IPN~Hand. The average $\xi_t$ peak levels reach $6.2$ at the beginning of each gesture versus a background level of $4.3$, with rapid changes between gestures. Therefore, the $90$\,\% recall operating point is set based on the $\tau$ value representing the $82$nd percentile.}
  \label{fig:xi}
\end{figure}

\textbf{Modality routing.} Per-group residual generators are computed for each of the three types of hand joint groupings (palm, fingertips, full hand), and the router selects the modality with $\arg\min_k \xi_t^{(k)}$. Based upon the $30{,}000$ frames collected for IPN~Hand, the router achieves $P_D = 0.722 \pm 0.005$ at $30.6$\,\% of the total ensemble MACs. At this percentage level, there is a reduction of $69.4$\,\% in per-frame energy consumption. Additionally, this performance level lies within $\pm 0.005$ of a supervised oracle gate. Using a reduced complexity feed-forward gate fit to the full generator, there is a drop in routing latency of $2.47\times$ at a cost of $0.003$ in terms of $P_D$.

\textbf{Drift-aware tracking.} Using the EKF measurement noise re-estimation block, the RMS tracking error is reduced by $4.6\times$ at $b_s = 5$\,cm when compared to the adaptive Kalman noise re-estimation baseline of~\cite{kruse2025adaptive}, and by $2\times$ when compared to the hybrid IMM-style filter of~\cite{or2022hybrid}. When averaged over all possible combinations of physics and bias parameters contained in the $27 \times 27$ grid defined by these parameters, the median ratio between the RMS error achieved by the calibrated gate and the RMS error achieved by a camera-only reference is $0.39$. There are three startup scalars (velocity autocorrelation, smoothness, boundedness ratio over the first second of operation) that determine which of two pre-tuned $\xi$-gate configurations to use, without using per-trace labels.

\begin{figure}[H]
  \centering
  \includegraphics[width=0.92\linewidth]{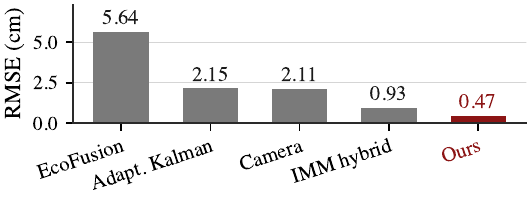}
  \caption{Tracking bias drift at $b_s = 5$\,cm. The EKF measurement noise re-estimation block reduces tracking RMS by $4.6\times$ when compared to the adaptive Kalman noise re-estimation baseline of~\cite{kruse2025adaptive} and by $2\times$ when compared to the hybrid IMM-style filter of~\cite{or2022hybrid}.}
  \label{fig:bias}
\end{figure}

\textbf{Comparison.} Table~\ref{tab:cmp} compares the number of MACs used per MCU and the memory required per MCU for the proposed system with those of two recently measured embedded radar front ends and one large convolutional baseline. The shared-residual architecture requires roughly four to fifty times fewer MACs than several recent measured embedded gesture pipelines~\cite{scherer2021tinyradar,zhang2024transcnn}; additionally, this architecture fits entirely within $14.4$\,KB of program memory on a single Cortex-M4F class part, leaving space for the EKF and router to reside on-chip. We explicitly refrain from asserting a power efficiency advantage; this will be evaluated during fabricated silicon measurements.

\begin{table}[H]
\centering
\caption{Comparison of wearable front ends utilizing mmWave and IMU. \textsf{W}: wake-up, \textsf{R}: routing, \textsf{D}: drift tracking. Baseline systems have measured silicon for wake-up alone; this paper provides a design study combining all three features with a single residual.}
\label{tab:cmp}
\footnotesize
\setlength{\tabcolsep}{4pt}
\begin{tabular}{l r r c c c}
\toprule
Front end & MACs/fr. & Memory & W & R & D \\
\midrule
\textbf{\textit{This work}} & \textbf{$110$\,K} & \textbf{$14.4$\,KB} & $\bullet$ & $\bullet$ & $\bullet$ \\
Scherer~\cite{scherer2021tinyradar} & $\sim$500\,K & $\sim$92\,KB & $\bullet$ & -- & -- \\
Zhang~\cite{zhang2024transcnn} & $\sim$5\,M & n/a & $\bullet$ & -- & -- \\
Safa~\cite{safa2021snn} & spiking & n/a & $\bullet$ & -- & -- \\
\bottomrule
\end{tabular}
\end{table}

\section{Failure Modes and Operating Envelope}
\label{sec:failures}

Shared-residual architectures have three operational envelopes that future hardware studies must consider. \emph{An adversary corrupting data after the selected modality is determined.} After determining a selected modality and selecting such modality, an adversary able to observe the output of the wake-up and inject additional noise with energy greater than five times that of the nominal noise floor onto that modality will defeat reactive selection, since $\xi_t$ rises on the corrupted channel only after that channel has been selected. Non-adaptive corruption was considered in this research; therefore, on-body deployment will likely require either a randomization of router fallback or a separate adversary detection head that exists outside this operational envelope.

\emph{Bias drift faster than the EWMA window.} The EKF measurement noise re-estimation tracks bias up to frequencies of order $1/(N\alpha)$, where $N$ represents the EWMA window length and $\alpha$ represents the smoothing parameter. Faster bias oscillations reduce the diagnostic signals provided by $\xi_t$; therefore, on the controlled dynamics benchmark we measure an RMS ratio degradation toward the camera-only reference when the bias cycle time falls below five EWMA windows. The values for $N$ and $\alpha$ used in calibration are stored in flash and may not be dynamically changed during deployment.

\emph{Heavy-tailed measurement noise.} Innovations with Student-$t$ distributions having degrees of freedom close to $3$ increase $R_k$ globally rather than locally to the transducers, thereby suppressing routing signals across all sensors and forcing the system toward its camera-only fallback. A Huber-clipped innovation norm is an obvious remedy~\cite{kruse2025adaptive} but is reserved for future work. As presented here, no extrapolation beyond what was seen is made regarding the above operational envelopes for the IPN~Hand, SHREC~2021, MiliPoint, and EAT-Radar datasets.

\section{Conclusion}
\label{sec:concl}

This appears to be the first design study to provide a wearable wristband front end using mmWave and IMU powered solely via a coin cell battery that includes wake-up gates, modality routers, and EKF measurement reweighting blocks generated by a single on-chip residual generator of size $278$\,B. The full system achieves $P_D = 0.72/0.80$ at $P_\text{FA} = 0.01$ with an average $47$\,\% reduction in classifier invocation energy per frame at a recall rate of $90$\,\%. Future work involves measuring silicon power on an Apollo4 Blue Plus board, collecting synchronized on-body capture datasets using IRB protocol, and analyzing the characteristics of the integrated FMCW radar (NF, SNR, range resolution, LoD).

\balance
\bibliographystyle{IEEEtran}
\let\OLDthebibliography\thebibliography
\renewcommand{\thebibliography}[1]{%
  \OLDthebibliography{#1}%
  \fontsize{6.5pt}{7.4pt}\selectfont
  \setlength{\itemsep}{0pt plus 0.2pt}%
  \setlength{\parsep}{0pt}%
}
\bibliography{refs}

\end{document}